\documentclass[runningheads,orivec]{llncs}
\usepackage{xcolor}
\usepackage{dsfont}
\usepackage{url}
\usepackage{subcaption}
\usepackage{graphicx} 
\usepackage{float}
\usepackage[T1]{fontenc}
\usepackage{csquotes}
\usepackage{hyperref}
\usepackage{graphicx}
\usepackage{amsfonts}
\usepackage{amsmath, amssymb}
\usepackage{tikz}
\makeatletter
\renewcommand\paragraph{\@startsection{paragraph}{4}{\z@}%
  {1.8ex \@plus1ex \@minus.3ex}%
  {-1em}%
  {\normalfont\normalsize\bfseries}}
\makeatother

\MakeOuterQuote{"}
\begin{document}
\usetikzlibrary{shapes, arrows.meta, positioning, shadows.blur, backgrounds, fit, calc}

\title{From Model-Based Screening to Data-Driven Surrogates: A Multi-Stage Workflow for Exploring Stochastic Agent-Based Models}
\titlerunning{Automated Exploration of Parameters for Agent-Based Simulation}
%
\author{Paul Saves\inst{1}\orcidID{0000-0001-5889-2302} 
\and Matthieu Mastio\inst{1}\orcidID{0009-0002-3486-3739} 
\and Nicolas Verstaevel\inst{1}\orcidID{0000-0002-7879-6681}
\and Benoit Gaudou\inst{1}\orcidID{0000-0002-9005-3004}
}
\authorrunning{P. Saves, M. Mastio, N. Verstaevel, B. Gaudou}
%
\institute{IRIT, Université Toulouse Capitole, Toulouse, France. 
\email{\{first.last\}@irit.fr}
%
\maketitle              
\begin{abstract}
The systematic exploration of Agent-Based Models (ABMs) is challenged by the curse of dimensionality and their inherent stochasticity.
We present a multi-stage pipeline integrating the systematic design of experiments with machine learning surrogates. Using a predator-prey case study, our methodology is carried out in two steps. First, an automated model-based screening identifies dominant variables, assesses outcome variability, and segments the parameter space. Second, we train Machine Learning models to map the remaining nonlinear interaction effects. This approach automates the discovery of \textit{unstable} regions where system outcomes are highly dependent on nonlinear interactions between many variables. Thus, this work provides modelers with a rigorous, hands-off framework for sensitivity analysis and policy testing, even when dealing with high-dimensional stochastic simulators.
\end{abstract}

\section{Introduction}
Agent-based models are powerful tools for simulating complex systems based on autonomous agents that interact within an environment. 
These models are particularly important for studying emergent phenomena when local interactions give rise to global behaviors that are often difficult to capture with traditional analytical methods~\cite{debosscher2023}.
In ecology and socio-environmental systems, ABMs are particularly appealing because they allow researchers to relax strong analytical assumptions (\textit{e.g.}, well-mixed populations or perfect rationality) and explicitly represent space, stochasticity, and interaction networks~\cite{banos2015agent}. Despite their flexibility, ABMs are often criticized for lacking systematic exploration protocols, making it difficult to assess the robustness, generality, and policy relevance of their results~\cite{railsback2019agent}.

In this paper, we propose a general protocol for model exploration that systematically investigates the behavior of agent-based simulation models as a function of actionable parameters.
Rather than focusing on empirical calibration or prediction, our objective is to support exploratory analysis and \emph{what-if} reasoning. The proposed protocol is designed to be transferable across simulation models for which external levers, such as regulatory, legal, or institutional constraints, can be meaningfully applied.
To illustrate this protocol, we rely on a deliberately simple toy model inspired by predator–prey dynamics with renewable resources~\cite{grimm2013individual}. 
We aim to enhance the credibility of model predictions by systematically quantifying their sensitivity to parameter variability and intrinsic stochasticity. To support reproducibility, the open-source research pipeline, including the NetLogo simulation model~\cite{wilensky1999netlogo,novak2011bughunt}, analysis scripts, datasets, and supplementary figures, is freely available in the online repository\footnote{\color{blue}{\url{https://github.com/ANR-MIMICO/MABS2026_Preys_Predators}}}.

The remainder of this paper is organized as follows. Section~\ref{sec:sota} positions our work within the existing literature, and Section~\ref{sec:model} details the spatial predator-prey simulation used as a case study. Section~\ref{sec:pipeline} introduces our proposed multi-stage workflow: Section~\ref{sec:doe} presents the experimental design and quantifies the simulation's intrinsic stochasticity; Section~\ref{sec:model_based} performs a preliminary model-based screening using linear and tree-based methods; and Section~\ref{sec:data_based} details the machine learning surrogate approach for nonlinear sensitivity analysis and uncertainty quantification. Finally, Section~\ref{sec:conclusion} summarizes our contributions and discusses future perspectives.

\section{Related works}
\label{sec:sota}

Assessing the robustness and validity of Agent-Based Models (ABM) requires moving beyond single-trajectory simulations toward systematic global exploration.
Still, the combination of high-dimensional parameter spaces, nonlinear dynamics, and intrinsic stochasticity makes this task computationally prohibitive~\cite{banos2015agent}.
In complex systems analysis, the primary goal of model exploration is to map the relationship between input parameters and output variability. While early approaches relied on local one-at-a-time methods, the field has shifted toward Global Sensitivity Analysis (GSA) to capture interactions over the full parameter space~\cite{iooss2015review}. Variance-based methods are often considered the gold standard for GSA as they decompose the output variance into contributions from single parameters and their interactions. However, calculating these indices requires Monte-Carlo sampling schemes that are computationally expensive for slow-running ABMs. Furthermore, simpler screening methods may fail to capture the complex, non-monotonic responses typical of ecological or social simulations, potentially leading to misleading conclusions about parameter importance~\cite{thiele2014facilitating}.

To tackle the computational cost associated wth GSA, one can use surrogate models for data augmentation. They generally consist of statistical or machine learning approximations trained on a limited set of ABM simulations to predict outcomes at a fraction of the cost~\cite{angione2022}. To do so, Gaussian processes or random forests have historically been preferred for their built-in uncertainty estimates, but they struggle to scale beyond low-dimensional spaces. Consequently, more complex methods such as gradient boosting, as well as deep neural networks, have gained traction~\cite{angione2022}. The latter may be less interpretable but offers the flexibility to approximate highly nonlinear response surfaces, interactions, and discontinuous regimes (\textit{e.g.}, tipping points) inherent to multi-agent systems~\cite{lamperti2019towards}.

A critical yet often overlooked challenge in ABM exploration is the distinction between sources of uncertainty. Standard surrogate approaches often treat the simulator as a deterministic blackbox, ignoring the \textit{aleatoric uncertainty} (intrinsic stochasticity) arising from random seeds in the ABM. Conversely, \textit{epistemic uncertainty} arises from the surrogate's lack of training data in certain regions. Failing to separate these can lead to overconfident predictions in unstable regimes~\cite{hullermeier2021aleatoric}.
Furthermore, the blackbox nature of complex machine learning surrogate models poses an interpretability challenge. In this context, conformal prediction has emerged as a critical framework for reliable uncertainty quantification. Unlike traditional Bayesian or error-variance methods, it provides distribution-free and finite-sample tools to construct rigorous prediction intervals that bound the surrogate's approximation error with a pre-specified confidence level~\cite{shafer2008tutorial}.
To bridge the gap between accurate prediction and mechanistic understanding, post-hoc explainable techniques such as, for example, Partial Dependence Plots (PDP) and Individual Conditional Expectation (ICE) are becoming essential~\cite{goldstein2015peeking}. These tools allow modelers to visualize not just which parameters matter, but how they influence the system (\textit{e.g.}, identifying thresholds and phase transitions). In this paper, we propose a pipeline that unifies these elements: leveraging machine learning whenever necessary for nonlinear approximations while rigorously decomposing uncertainty to identify robust stability boundaries and critical regimes.

\section{Model Description}
\label{sec:model}
\paragraph{Baseline Model.}
We extend a classical toy predator–prey model~\cite{novak2011bughunt} by introducing three spatially explicit mechanisms that drive complex system dynamics.  
Two types of agents interact in this model: herbivores (bandicoots) gain energy by consuming the renewable resource, while predators (foxes) gain energy by consuming herbivores. The simulation is implemented in Netlogo~\cite{banos2015agent,wilensky1999netlogo} and is stopped after $1000$ timesteps ($t$)~\footnote{\color{blue}{\url{http://ccl.northwestern.edu/netlogo/}}}.
The environment is defined as a square lattice of $60 \times 60$ discrete patches. The simulation environment is defined as a toroidal grid to eliminate edge effects and avoid boundary handling.

Each patch contains a renewable resource (grass) that serves as the primary energy source for herbivorous agents. 
Resource availability on patch $i$ at time $t$ is represented by a continuous variable $R_i(t) \in [0, R_{\max}]$.
Grass growth follows the  regenerative process as \(
R_i(t+1) = \min \left( R_{\max},\; R_i(t) + g \right)
\),  where $g$ is the intrinsic growth rate of the resource. When grass is consumed, $R_i(t)$ is reduced proportionally to the number of herbivores on the patch and their intake.
Each agent $a$ of species $s$ is characterized by its position $\mathbf{x}_a(t)$, its energy $E_a(t)$, its age $A_a(t)$ and its maximum lifespan $A_a^{\max}$. Native herbivores consume grass in their current patch. When sufficient resources are available, grass energy is reduced. Similarly, predators feed on native herbivores occupying the same patch, instantly removing a prey. In both cases, the successful agent gains energy according to 
\( E_a(t+1) = E_a(t) + \alpha_s, \)
where $\alpha_s$ is the energy gain parameter specific to the species $s$.
Energy decreases by 1 every tick and increases through successful feeding.
Agents die if their energy becomes negative (\(E_a(t) < 0\))  or if their age exceeds the species-specific maximum lifespan (\(A_a(t) > A_a^{\max} \)).

Reproduction is asexual and energy-dependent. An agent has 50\% chance of reproducing if
\(
E_a(t) \geq E_s^{\text{rep}} \quad \text{and} \quad A_a(t) \geq A_s^{\text{rep}},
\) 
where $E_s^{\text{rep}}$ and $A_s^{\text{rep}}$ are species-specific thresholds.
The number of offspring is drawn from a bounded discrete distribution, but since each child receives $E_s^{\text{rep}}$ energy from its parent, the maximum number of offspring is therefore proportional to the parent's energy. Offsprings spawn one stride distance away from their parents.
\paragraph{Spatial extensions.}
Unlike classical predator--prey models assuming homogeneous resource availability~\cite{novak2011bughunt}, grass is distributed non-uniformly across space. Initial grass patches are generated around a limited number of spatial centers, with the probability that a patch is fertile decreasing exponentially with its distance to the nearest center:
\(\mathbb{P}(i \text{ is fertile}) \propto \exp(-k d_i),
\)
where $d_i$ denotes the distance from patch $i$ to the closest grass cluster center and $k$ controls the spatial decay rate. This mechanism produces clustered resource landscapes, introducing spatial heterogeneity and localized competition.

Also, compared to the dummy implementation, agent movements are not purely stochastic when food sources or prey are locally available. Instead, agents exhibit directed movement toward relevant targets within their perceptual range to improve the spatialization of the model and the interactions between the agents and their environments.
Each agent of species $s$ is endowed with a perception radius $r_s$, defining a local neighborhood $\mathcal{N}_a$ within which relevant environmental information can be detected.
Herbivorous agents move toward the nearest patch containing available grass within $\mathcal{N}_a$. Predatory agents move toward the nearest detectable prey individual within $\mathcal{N}_a$.
When no suitable target is detected within the perception neighborhood, agents default to a random walk. 

Finally, we introduce a spatial risk component through the implementation of localized hunting zones. A subset of the fertile patches is randomly designated as regulated hunting areas. In these locations, both native herbivores and predators are subject to an additional stochastic death probability, $P_{death, s}$, which is proportional to the established hunting quotas for species $s$. This mechanism simulates anthropogenic pressure and introduces a non-biological source of mortality that varies across the landscape.

\section{Multi-Stage Exploration Pipeline}
\label{sec:pipeline}

The primary challenge in exploring stochastic ABMs lies in the computational cost of resolving high-dimensional interactions while accounting for aleatoric noise. To address this, we propose a hierarchical ``zoom-in'' pipeline, illustrated in Fig.~\ref{fig:workflow}, designed to bridge the gap between global statistical trends and local mechanical realities. 

The methodology transitions from a coarse \textit{model-based screening} to a fine-grained \textit{data-driven analysis}. Initially, we assess internal stochasticity using linear and simple global screening to identify dominant drivers and interpretable extraction rules methods to segment stability regions locally. We then deepen this analysis by training, if needed, one or many machine learning surrogate models to capture non-linear dynamics. This second stage facilitates a rigorous dual assessment: variance-based global sensitivity analysis ranks interactions~\cite{iooss2015review}, while local explainability tools map the precise functional forms and tipping points governing ecosystem resilience~\cite{mastio2026adaptive}. This paper illustrates a two-stage workflow, but depending on the underlying model at hand, more steps may be required, or the first step could be sufficient. 

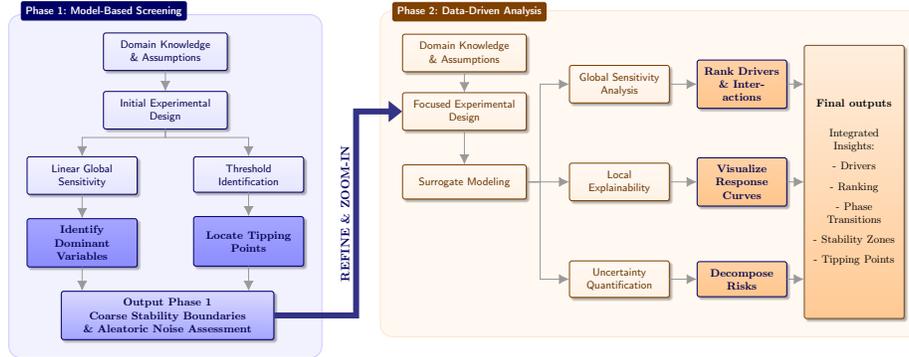
\begin{figure}[t!] 
\centering 

\resizebox{\textwidth}{!}{%
\begin{tikzpicture}[
    font=\sffamily\footnotesize,
    >={LaTeX[width=3mm,length=3mm]},
    node distance=0.8cm and 0.6cm,
    box/.style={
        rectangle, 
        draw, 
        rounded corners=2pt, 
        align=center, 
        text width=3.2cm, 
        minimum height=1cm,
        inner sep=6pt,
        blur shadow={shadow blur steps=5}
    },
    boxP1/.style={
        box,
        draw=blue!40!black,
        top color=blue!5,
        bottom color=blue!10,
        text=blue!30!black
    },
    titleP1/.style={
        fill=blue!40!black,
        text=white,
        font=\bfseries\sffamily,
        rounded corners=2pt,
        inner sep=4pt
    },
    boxP2/.style={
        box,
        draw=orange!50!black,
        top color=orange!5,
        bottom color=orange!10,
        text=orange!40!black
    },
    titleP2/.style={
        fill=orange!50!black,
        text=white,
        font=\bfseries\sffamily,
        rounded corners=2pt,
        inner sep=4pt
    },
    boxFinal/.style={
        box,
        text width=2.5cm,
        minimum height=6cm,
        draw=brown!60!black,
        top color=orange!13,
        bottom color=orange!42,
        font=\bfseries
    },
    boxP1action/.style={
    boxP1,
    fill=blue!135,
    font=\bfseries,
    top color=blue!35,
    bottom color=blue!45,
    },
    boxP1f/.style={
    boxP1,
    fill=blue!135,
    font=\bfseries,
    top color=blue!15,
    bottom color=blue!35,
    },
    boxP2action/.style={
    boxP1,
    fill=orange!135,
    font=\bfseries,
    top color=orange!35,
    bottom color=orange!45,
    },
    line/.style={draw, thick, ->, color=gray!80},
    thickarrow/.style={draw, line width=1.75mm, -{Triangle[length=4mm, width=6mm]}, color=blue!40!black!80}
]

    
    \node[boxP1] (input) {Domain Knowledge\\\& Assumptions};
    
    \node[boxP1, below=0.6cm of input] (doe1) {Initial Experimental\\Design};
    
    \node[boxP1, below left=0.8cm and -1cm of doe1, text width=2.8cm] (anova) {Linear Global Sensitivity};
    \node[boxP1, below right=0.8cm and -1cm of doe1, text width=2.8cm] (tree) {Threshold Identification};
    
    \node[boxP1action, below=0.7cm of anova, text width=2.8cm, fill=blue!30!cyan!10] (dom) {Identify Dominant\\Variables};
    \node[boxP1action, below=0.7cm of tree, text width=2.8cm, fill=blue!130!cyan!10, minimum height = 1.45cm] (tip) {Locate Tipping\\Points};
    
    \node[boxP1f, below=0.7cm of dom, text width=5.8cm, xshift=2.5cm] (out1) {\textbf{Output Phase 1}\\Coarse Stability Boundaries \& Aleatoric Noise Assessment};

    \draw[line] (input) -- (doe1);
    \draw[line] (doe1) -- ++(0,-0.8) -| (anova);
    \draw[line] (doe1) -- ++(0,-0.8) -| (tree);
    \draw[line] (anova) -- (dom);
    \draw[line] (tree) -- (tip);
    \draw[line] (dom) -- (out1.north -| dom);
    \draw[line] (tip) -- (out1.north -| tip);

    \node (phase2NW) [right=5cm of input.north east] {};

    \node[boxP2,below=1.75cm of phase2NW,  anchor=north west] at (phase2NW)
    (doe2) {Focused Experimental\\Design};
    
    \node[boxP2, above =0.6cm of doe2 ] (input2) {Domain Knowledge\\\& Assumptions};

    \node[boxP2, below=1cm of doe2] (surr)
    {Surrogate Modeling};

    \node[boxP2, right=1.25cm of surr, text width=2.5cm] (pdp) {Local\\ Explainability};
    \node[boxP2, above=1.75cm of pdp, text width=2.5cm] (sobol) {Global Sensitivity\\Analysis};
    \node[boxP2, below=1.75cm of pdp, text width=2.5cm] (uq) {Uncertainty\\Quantification};
    
    \node[boxP2action, right=0.8cm of sobol, text width=2.2cm, fill=orange!20] (act1) {Rank Drivers\\\& Interactions};
    \node[boxP2action, right=0.8cm of pdp, text width=2.2cm, fill=orange!20] (act2) {Visualize\\Response Curves};
    \node[boxP2action, right=0.8cm of uq, text width=2.2cm, fill=orange!20] (act3) {Decompose\\Risks };

    \node[boxFinal, right=0.5cm of act2, minimum width = 2.6cm, minimum height=8cm ] (final) {Final outputs\\[0.5cm]\normalfont\footnotesize Integrated Insights:\\[0.2cm]- Drivers\\[0.2cm] - Ranking\\[0.2cm] - Phase Transitions\\[0.2cm]- Stability Zones\\[0.2cm] - Tipping Points };

        \draw[line] (input2) -- (doe2);
    \draw[line] (doe2) -- (surr);
    \draw[line] (surr) -- (pdp);
    \draw[line] (surr.east) -- ++(0.4,0) |- (sobol);
    \draw[line] (surr.east) -- ++(0.4,0) |- (uq);
    \draw[line] (sobol) -- (act1);
    \draw[line] (pdp) -- (act2);
    \draw[line] (uq) -- (act3);
    \draw[line] (act1.east) -- ++(0.5cm, 0);
    \draw[line] (act2) -- (final);
    \draw[line] (act3.east) -- ++(0.5cm, 0);
    \draw[thickarrow] (out1.east) -- ++(2.4,0) |- (doe2.west) node[pos=0.25, rotate=90, above, font=\bfseries\normalsize, text=blue!40!black] {REFINE \& ZOOM-IN};

    \begin{scope}[on background layer]
        \node[fit=(input)(out1)(anova)(tree), fill=blue!5, draw=blue!20, rounded corners=10pt, inner sep=15pt] (bg1) {};
        \node[titleP1, above right=0cm and 0cm of bg1.north west, anchor=west, xshift=10pt, yshift=3pt] {Phase 1: Model-Based Screening};
        
        \node[fit=(phase2NW)(doe2)(final)(uq)(sobol), fill=orange!5, draw=orange!20, rounded corners=10pt, inner sep=15pt] (bg2) {};
        \node[titleP2, above right=0cm and 0cm of bg2.north west, anchor=west, xshift=10pt, yshift=-1pt] {Phase 2: Data-Driven Analysis};
    \end{scope}

\end{tikzpicture}
}
\caption{{Multi-stage exploration workflow.}}
\label{fig:workflow} 
\vspace{-0.5cm}
\end{figure}

\paragraph{Domain Knowledge and Assumptions}

\label{sec:doe}

To systematically explore the model's behavior and the interactions between spatial heterogeneity and agent behavior, we identify 13 continuous variables and one categorical seed variable. These parameters characterize environmental constraints, prey (bandicoot) metabolism, and predator (fox) efficiency. The experimental variables and their respective ranges are summarized in Table \ref{tab:parameters}.
We assumed a uniform distribution across the 13-dimensional input space, $E$ denotes energy units, and $t$ denotes ticks. 
\begin{table}[b]
\vspace{-0.8cm}
\setlength{\arrayrulewidth}{1pt} 
\renewcommand{\arraystretch}{1.2} 
\centering
\scriptsize
\caption{Model parameters, symbols, and experimental ranges.}
\label{tab:parameters}
\begin{tabular}{llcl}
\hline
\textbf{Category} & \textbf{Parameter} & \textbf{Symbol} &  $\  $ \textbf{Range / Units} \\ \hline
\textit{Environment} & Grassland proportion on the map & $Gr$ & $[10.0, 100.0]\, (\%)$ \\
  & Proportion of hunting zone among grassland & $PH$ & $[0.0, 100.0]\, (\%)$ \\
 \hline
\textit{Plants}
 & Maximum Plant Energy & $PM$ & $[50.0, 250.0]\, (E)$ \\
  & Plant Growth Rate & $PR$ & $[5.0, 25.0]\, (E/t)$ \\
 & Grass energy intake of eating bandicoots & $BF$ & $[2.0, 10.0]\, (E)$ \\
  \hline
\textit{Bandicoots}  & Energy Gain when eating & $BG$ & $[1.0, 20.0]\, (E)$ \\
 & Energy reserves required for reproduction & $BR$ & $[8.0, 20.0]\, (E)$ \\
 & Hunting quota on hunting zone & $BH$ & $[0.0, 100.0]\, (\%)$ \\
 & View Radius to search for grass& $BV$ & $[0.0, 3.0]\, (\text{cells})$ \\ \hline
\textit{Foxes}  & Energy Gain when eating  & $FG$ & $[10.0, 50.0]\, (E)$ \\
  & Energy reserves required for reproduction  & $FR$ & $[12.0, 30.0]\, (E)$ \\
  & Hunting quota on hunting zones & $FH$ & $[0.0, 100.0]\, (\%)$ \\
 & View Radius for hunting bandicoots & $FV$ & $[0.0, 3.0]\, (\text{cells})$ \\ \hline

\textit{Replications} & Random Seed & $S$ & $\{1, 2, 3, 4, 5\}$ \\ \hline
\end{tabular}
\end{table}

To illustrate our methodology, we focus on population subsistence as the primary model output, $y \in \{\texttt{extinction}, \texttt{prey\_survival}, \texttt{coexistence}\}$ that is an ordinal variable~\cite{saves2026modeling}. For computational simplicity, we encode this output numerically as $y \in \{0, 0.5, 1\}$. Although assigning $0.5$ to the intermediate state is an approximation and sophisticated methods exist to learn nonlinear warping operators for ordinal data~\cite{saves2026modeling}, we adopt this linear encoding as a convenient proxy for stability ordering. This simplification does not impact the classification-based surrogate analyses performed later in the study.

\subsection{Model-based Analysis}
\label{sec:model_based}
Following our workflow, we analyze the raw simulation data to identify global trends and assess the impact of stochasticity. This step informs the subsequent data-driven surrogate modeling by highlighting the limitations of linear assumptions. We first start with a global analysis, then we dive into more local insights.

\paragraph{Initial Experimental Design and Data Analysis}

To ensure efficient and space-filling sparse exploration, we employed {Latin Hypercube Sampling (LHS)} to generate $N = 650$ unique parameter configurations ($50$ points times the number of variables). 
To account for model stochasticity, each configuration was executed across $n=5$ independent replications using distinct random seeds, resulting in a total of \textbf{3,250 simulation runs}.

In the data, the outcomes are significantly unbalanced because $60 \%$ of the data correspond to a total extinction, 27\% correspond to a prey survival, and only 13\% of the data correspond to a sustainable coexistence between the two species. Therefore, our first analysis will focus on understanding globally the response across the whole search space to identify the reason for such a low proportion of coexistence. Then, a second analysis will focus on understanding the drivers of coexistence.

We quantify the inherent stochasticity of the simulation to establish a theoretical performance benchmark. Since the model output $y$ is a random variable conditional on the seed $S$, no deterministic predictor can achieve $100\%$ accuracy. We explicitly measured this \textit{Aleatoric Uncertainty} to distinguish between surrogate modeling error and irreducible simulation noise (Bayes error rate). To determine the upper bound of predictability, we constructed a theoretical oracle predictor consisting of the Bayes optimal classifier~\cite{tumer1996estimating}. For each unique parameter configuration $X_i$, we aggregated the five replicates $Y_i = \{y_{i,1}, \dots, y_{i,5}\}$ and computed the group median $\tilde{y}_i$. We effectively asked: \textit{"If a model perfectly learned the central tendency of the system, how often would it correctly predict the individual simulation runs?"}
In our model, the theoretical maximum accuracy ($Acc_{max}$) is defined as the proportion of individual runs that align exactly with their group median:
\( Acc_{max} = \frac{1}{N \times n} \sum_{i=1}^{N} \sum_{j=1}^{n} \mathbb{I}(y_{i,j} = \tilde{y}_i) \).
The analysis reveals a theoretical accuracy limit of {95.3\%}. This implies that only {4.7\%} of the variance is purely aleatoric (noise) and cannot be explained by the input parameters $X$.

\paragraph{Leveraging ANOVA for Linear Global Sensitivity Analysis.}

To quantify the contribution of each parameter to the variance of the population outcome ($y$), we conducted a simple ANOVA on a generalized linear model fitted by ordinary least squares. The ANOVA used Type II sums of squares to estimate the proportion of variance explained by each factor while controlling for other main effects~\cite{iooss2015review}.  Note that we could have opted for other methods, such as one-at-a-time Morris elementary effects~\cite{iooss2015review}.
Analyzing the results reveal a clear hierarchy of linear drivers. The Proportion of Hunting Zones ($PH$) is the dominant factor, explaining approximately 25.7\% of the total variance. This is followed by the Bandicoot Hunt Chance ($BH$, 9\%) and the Bandicoot Energy Gain when eating ($BG$, 7E). Within the linear approximation, these results suggest that anthropogenic pressure contributes more to variability in outcomes than most individual metabolic parameters. 
The Seed ($S$) factor explains negligible variance, implying that internal stochasticity does not bias global trends. A $\chi^2$ test across seeds ($\chi^2 = 1.25, p > 0.99$) further confirms that population regimes are independent of random initializations.
However, the linear model achieves an $R^2$ of only 43\%, leaving a large residual variance (56.5\%). This substantial unexplained variance indicates that nonlinear interactions and higher-order effects play a major role in system dynamics and motivates the use of nonlinear surrogate methods in subsequent analysis.
While the linear analysis highlights a strong sensitivity to anthropogenic parameters, the significant residual variance confirms that linear methods are insufficient to capture the model's full complexity. This motivates the subsequent use of nonlinear surrogate modeling to better understand these interactions.

\paragraph{Leveraging a Decision Tree for Threshold Identification.}

To resolve the nonlinearities identified by the ANOVA residuals, we trained a regression tree 
to identify specific parameter thresholds that trigger regime shifts~\cite{chen2006decision}.  This allows for studying the identified local phenomena locally.
The tree identifies a primary tipping point at $PH \approx 31\%$. When hunting zone density exceeds this threshold, the system stability relies heavily on  $BH$. If $BH > 18\%$, the outcome converges toward total extinction (mean value of 8\%). Conversely, in low hunting zone environments, metabolic factors become critical due to the fact that there is less hunting zones. In that case, a low energy gain for when bandicoots are eating,  $BG < 4$, also leads toward the total extinction of both species (7\% on average).

\paragraph{First Refinement of the Search Space.}
To balance computational feasibility with statistical rigor, we determined the required number of replications, or number of seeds, ($n$) using a sequential procedure based on the asymptotic normality of the sample mean. Relying on the central limit theorem, we assume that for a sufficiently large number of replications, the sampling distribution of the mean stability score approximates a Gaussian distribution, regardless of the underlying population distribution \cite{montgomery2017design}.
We targeted a confidence level of $95\%$ ($Z_{\alpha/2} \approx 1.96$) with a desired margin of error of $\epsilon = \pm 0.1$. The optimal sample size $n^*$ for a given parameter configuration is estimated as \( n^* = \left( \frac{Z_{\alpha/2} \cdot \hat{\sigma}}{\epsilon} \right)^2  \), 
where $\hat{\sigma}$ is the sample standard deviation estimated from previous runs. Our analysis of the pilot data revealed that the average required sample size across the input space is ${n}^* = 20.58$. Consequently, we fixed the experimental design at $n=20$ replicates. This choice satisfies the statistical requirements for the majority of the parameter space. 

Parts of the parameter space yield trivial extinction outcomes. Therefore the subsequent search space exploration will restrict the operational range of $PH$ to $[0, 30]\%$, $BH$ to $[0, 20]\%$ and $BG$ to $[5,20]$, to focus computational resources on the transition zone where complex coexistence or bandicoots survival dynamics occur, especially since the ANOVA indicates that nonlinearities and interactions effects explains most of the variations in the simulation outcomes. Therefore, this first analysis successfully identified the search space regions leading to non-coexistence, and our second analysis can effectively focus on understanding the drivers of coexistence.

\subsection{Data-based analysis}
\label{sec:data_based}
Building on our preliminary findings, we sampled a new LHS of 650 points, each replicated 20 times for a total of 13,000 data points. The new class imbalance is of 16\% extinction, 26\% prey survival, and 58\% coexistence. 
The theoretical accuracy limit is now 89.6\%, implying a minimum of around 10\% randomness in the results~\cite{hullermeier2021aleatoric}. 
To capture non-linear phase transitions, we train an MLP classifier (2 $\times$ 128 neurons, ReLU) using $\mathcal{L}^2$ regularization to prevent overfitting the irreducible noise~\cite{goodfellow2016deep}. Validated via Stratified Group 10-Fold Cross-Validation, the surrogate achieves $80.1\%$ accuracy.
Error analysis indicates that misclassifications primarily occur at the metastable boundary between coexistence and predator extinction.

While the surrogate classifier is trained on three discrete states, the subsequent sensitivity and interpretability analyses focus specifically on the predicted probability of the coexistence regime, $P(y = \texttt{coexistence})$. This continuous metric will serve as a proxy for ecosystem resilience, allowing us to quantify how parameters drive the system toward or away from a steady state. Unlike discrete class labels, the probability score captures subtle gradients in stability, revealing regions where the ecosystem is technically stable but highly vulnerable to stochastic collapse versus regions of a more robust stability.

\paragraph{Leveraging Sobol' indices for nonlinear global sensitivity analysis.}

To quantify parameter influence, we contrast a linear ANOVA on ordinal regimes, performed on the new dataset, with a variance-based Sobol' analysis derived from an MLP surrogate (Table~\ref{tab:sensitivity_comparison}).
Note that, while the specific selection of surrogate architectures and GSA methods is not the primary focus of this study, the rationale and workflow for such choices are detailed in our previous work~\cite{saves2025surrogate}.
Sobol' indices decompose output variance to capture nonlinearities: the First-Order Index ($S_1$) measures a parameter's pure contribution, while the Total-Order Index ($S_T$) accounts for its full effect, including all interactions.  The two methods provide sharply different views of the input–output relationship: the linear ANOVA performs poorly (residual variance 87\%), indicating that system dynamics are largely governed by non-additive interactions and threshold effects beyond the reach of linear statistics. 

The comparison reveals a sharp disparity between linear and non-linear sensitivity assessments. The linear ANOVA model exhibits poor explanatory power (residual variance $>85\%$), suggesting a system dominated by top-down control where fox-related traits are the primary drivers. Under this linear assumption, only the direct effects of predation emerge above the noise, while all other parameters appear statistically insignificant ($<1\%$ explained variance).

In contrast, the Sobol' indices identify bandicoot-related parameters and environmental constraints as the dominant drivers. 
While predation is the proximal cause of mortality, coexistence is regulated bottom-up: stability depends on the prey's metabolic capacity and resource availability to buffer against stochastic extinction. 
The system nonlinearity is further characterized by the important gap between First-Order ($S_1$) and Total Order ($S_T$) effects. The low sum of first-order indices indicates that 69\% of the variance arises from higher-order parameter interactions, confirming that regime shifts are driven by complex synergies rather than isolated parameter effects. Every variable plays a role when considering interactions; $S_T$ varies from 46\% to 6.4\%. Note that the second-order Sobol' indices account for 55\% of the variance, and almost exclusively in interaction with the top 6 variables given in Table~\ref{tab:sensitivity_comparison}. Therefore, only 17\% of the variance comes from the interactions between 3 variables or more. Notably, the amount of food bandicoots eat ($BF$) is the only exception; it does not interact at order 2 but interacts strongly at high orders. This variable looks insignificant at orders 1 and 2, but plays an important role as a connecting factor between grassland-related variables and bandicoot-related variables.

\begin{table}[t]
\centering
\scriptsize
\setlength{\arrayrulewidth}{1pt} 
\renewcommand{\arraystretch}{1.1} 
\caption{Comparison of Sensitivity Metrics: ANOVA  vs. Sobol' indices ($S_1, S_T$).}\label{tab:sensitivity_comparison}
\scriptsize
\begin{tabular}{|l|c|c|c|}
\hline
\textbf{Parameter} 
& \textbf{\ ANOVA  (\%)\ } 
& \textbf{\ Sobol' $S_1$\ } 
& \textbf{Sobol' $S_T$}  \\
\hline
{Band. Energy Gain (BG)\ } 
& 0.09 
& \textbf{7.6} 
& \textbf{45.9}  \\

{Prop. Hunting (PH)} 
& 3.00 
& 4.7 
& 38.7 \\

{Fox Energy Gain (FG)} 
& 2.86 
& 2.2 
& 28.6  \\
{Grassland \% (Gr)} 
& 0.04 
& 5.2 
& 27.4  \\
{Fox Hunting (FH)} 
& \textbf{4.11} 
& 2.1 
& 21.8   \\

{Plant Growth Rate (PR)} 
& 0.85 
& 5.8
& 17.8 \\
\hline

\textit{Model Fit} 
& $R^2 = 12.9\%$ 
& \multicolumn{2}{c|}{\ Surrogate Accuracy $= 80.1\%\ $} \\
\hline

\end{tabular}
\end{table}

\paragraph{Leveraging PDP and ICE for Threshold Identification.}
To interpret the nonlinear effects highlighted by the Sobol' analysis, we computed Partial Dependence Plots (PDPs) and Individual Conditional Expectation (ICE) curves from the MLP surrogate’s predicted coexistence probability: PDPs show the marginal mean response, while ICE curves display individual \textit{ceteris paribus} deviations~\cite{goldstein2015peeking}.  
We then quantify uncertainty, using conformal Prediction implemented through a dual Random Forest scheme~\cite{shafer2008tutorial}. This approach disentangles the epistemic component \(\sigma_{\mathrm{epistemic}}\) (model uncertainty) from the aleatoric component \(\sigma_{\mathrm{aleatoric}}\) (intrinsic stochasticity). The components were combined, based on the additive nature of assumed independent variances, as
\(
\sigma_{\mathrm{total}}=\sqrt{\sigma_{\mathrm{aleatoric}}^{2}+\sigma_{\mathrm{epistemic}}^{2}}
\)~\cite{hullermeier2021aleatoric}.
Consequently, high values of \(\sigma_{\mathrm{total}}\) identify critical phase-transition zones. We show these dynamics over the six most important variables in Figure~\ref{fig:6fig}, which reveals critical thresholds and non-linear interactions described as follows.\\
\begin{figure}[t!]
    \centering
    \includegraphics[height=10cm, width=0.90\textwidth]{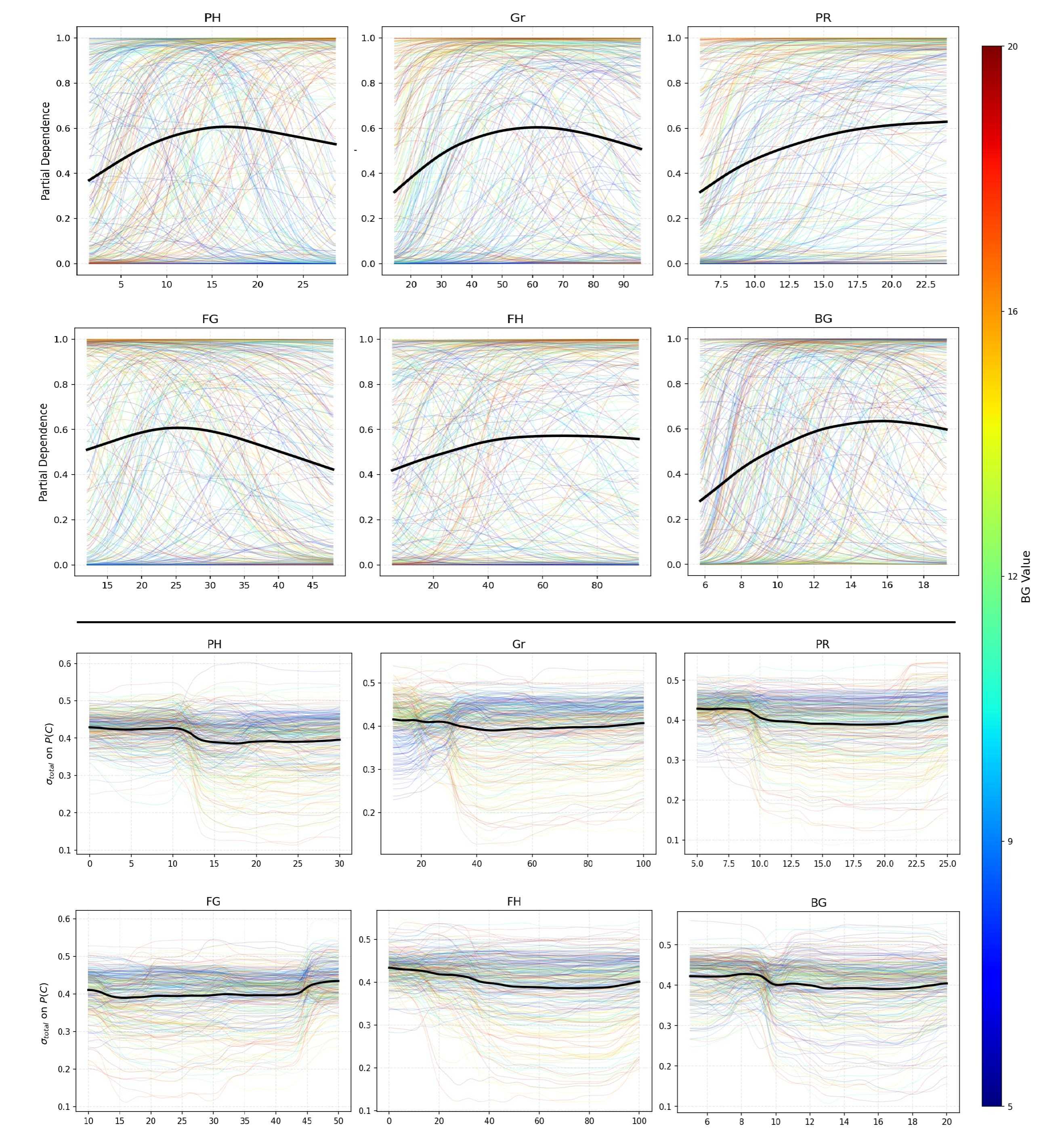}
    \caption{PDP/ICE and Uncertainties variations for the 6 most important features.}
    \label{fig:6fig}
    \vspace{-0.6cm}
\end{figure}
    $\bullet$ \textit{Metabolic Thresholds} (Bandicoots and Foxes energy gains $BG, FG$):
    $BG$ exerts a critical influence at low values. We observe a stabilization threshold between 10E and 15E; beyond this range, both the system dynamics and the associated uncertainty plateau.
    $FG$ exhibits minimal interaction with $BG$. It shows a non-monotonic effect on coexistence, peaking at $FG \approx 25$E before slowly declining. While overall uncertainty is stable, extreme values increase stochasticity. Specifically, the combination of low $FG$ and low $BG$ creates a highly unstable system, whereas excessive $FG$ leads to over-predation, driving the system directly from coexistence to total extinction.\\
    $\bullet$ \textit{Anthropogenic Paradox} (Probability of hunting and foxes quotas $PH$, $FH$):
    $PH$ promotes stability in robust metabolic regimes (high $BG$), stabilizing the system after a threshold of 10--15. However, under low metabolic conditions (low $BG$), increasing $PH$ reduces coexistence probability. This suggests that hunting is viable only in resilient ecosystems; in fragile environments, it accelerates collapse. A tipping point appears at $PH \approx 15\%$, marked by a sharp drop in uncertainty around $PH \approx 12\%$.
    Similarly, $FH$ favors coexistence up to a quota of $FH\approx 35\%$. Like $PH$, predator control is effective only when bandicoots are metabolically efficient (high $BG$). In these resilient environments, sufficient hunting minimizes uncertainty.\\
    $\bullet$ \textit{Spatial Resource Dynamics} (Amount of Grassland and Plant growth rate $Gr$, $PR$): 
    The interaction between $Gr$ and $BG$ explains $7\%$ of the total variance. While low values for both lead to deterministic extinction, high $BG$ accelerates convergence to coexistence as the grassland increases. Conversely, at low $BG$, excessive grassland favors a prey-only regime. The uncertainty landscape reveals a \textbf{regime shift} at $Gr \approx 28\%$: below this threshold, low $BG$ results in low uncertainty (deterministic extinction); above it, low $BG$ yields high uncertainty (outcome depends on stochastic fox survival), while high $BG$ yields low uncertainty (robust coexistence).
    $PR$ operates independently of $BG$. Higher growth rates correlate with increased stability and coexistence. A sharp drop in uncertainty occurs at a tipping point of $PR=9$E/t; below this, the system becomes highly sensitive to the stochastic spatial distribution of resources.

Quantifying both \textit{Total Uncertainty} and \textit{Partial dependence} 
is a methodological contribution that allow modelers to precisely identify tipping point regions where the ecosystem is structurally unstable, providing a rigorous framework to distinguish between deterministic regime shifts and irreducible stochastic risks.

\section{Conclusion and Perspectives}
\label{sec:conclusion}

In this work, we introduced a multi-stage, data-driven pipeline for the automated exploration of stochastic ABMs. By bridging the gap between classical Design of Experiments and Machine Learning surrogates, we addressed the dual challenge of high dimensionality and inherent stochasticity often found in complex simulations.
Our methodology, validated on a spatially explicit predator-prey simulation, demonstrates that linear methods, while useful for initial screening, fail to capture the critical non-linear metabolic interactions and threshold effects that govern ecosystem resilience.

Future works focus on three main axes. First, we will develop an active learning loop, where the uncertainty maps generated in the final stage are used as acquisition functions to iteratively refine the input space exploration with minimal additional simulations. Second, the differentiability of the trained surrogate model opens the way for gradient-based policy optimization, allowing for the automated discovery of robust management strategies in more complex socio-environmental digital twins. 
Finally, we advocate for a shift from manual, point-based calibration toward an automated, global characterization of the model's behavioral space. We recognize that the choice of a specific GSA method or surrogate architecture (e.g., ANOVA vs. Sobol, Random Forest vs. MLP) remains highly dependent on the model's structure and the problem at hand. Consequently, the ultimate goal is to connect this framework with recommendation tools within an \textit{AutoXAI} pipeline~\cite{saves2025surrogate}. This will automate the selection of the most appropriate analytical tools for a given problem, removing the bias of manual selection and standardizing the exploration of complex systems.

 \paragraph{Acknowledgements}
The research presented in this paper has been performed in the framework of the MIMICO research project funded by the Agence Nationale de la Recherche (ANR) n$^o$ ANR-24-CE23-0380.  This work was supported by the MUTTEC project (France 2030 – TIRIS, contract 23-AAP-TIRIS-01-062).

\bibliographystyle{splncs04}
\bibliography{main.bib}

\newpage
\appendix
\section*{Supplementary Materials: Generated Figures}

\label{sec:supplementary}


\begin{figure}[htbp]
    \centering
    \includegraphics[width=0.99\linewidth]{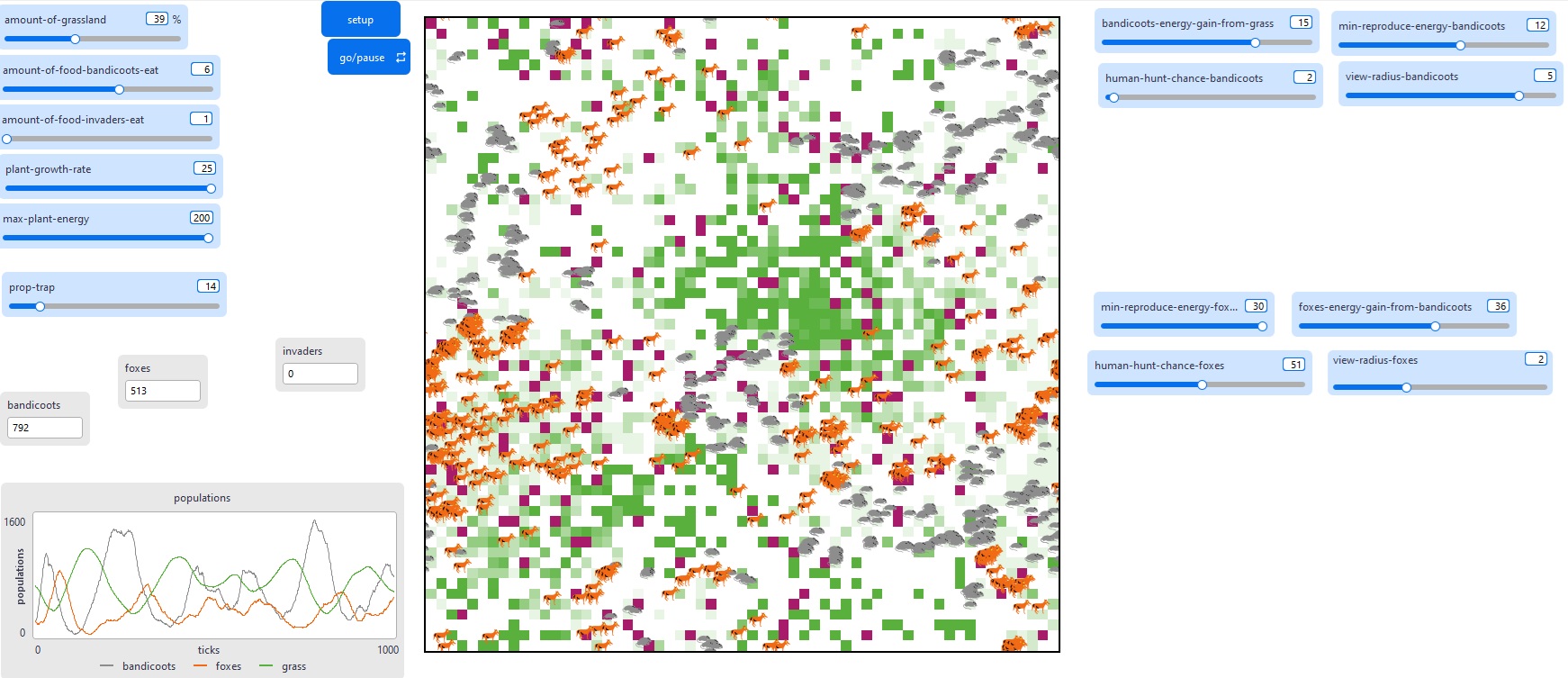}
    \caption{Simulation results showing the relationship between variables X and Y.}
    \label{fig:simulation_results}
\end{figure}

\begin{figure}[H]
    \centering
    \begin{subfigure}[b]{0.49\textwidth}
        \centering
        \includegraphics[width=\textwidth]{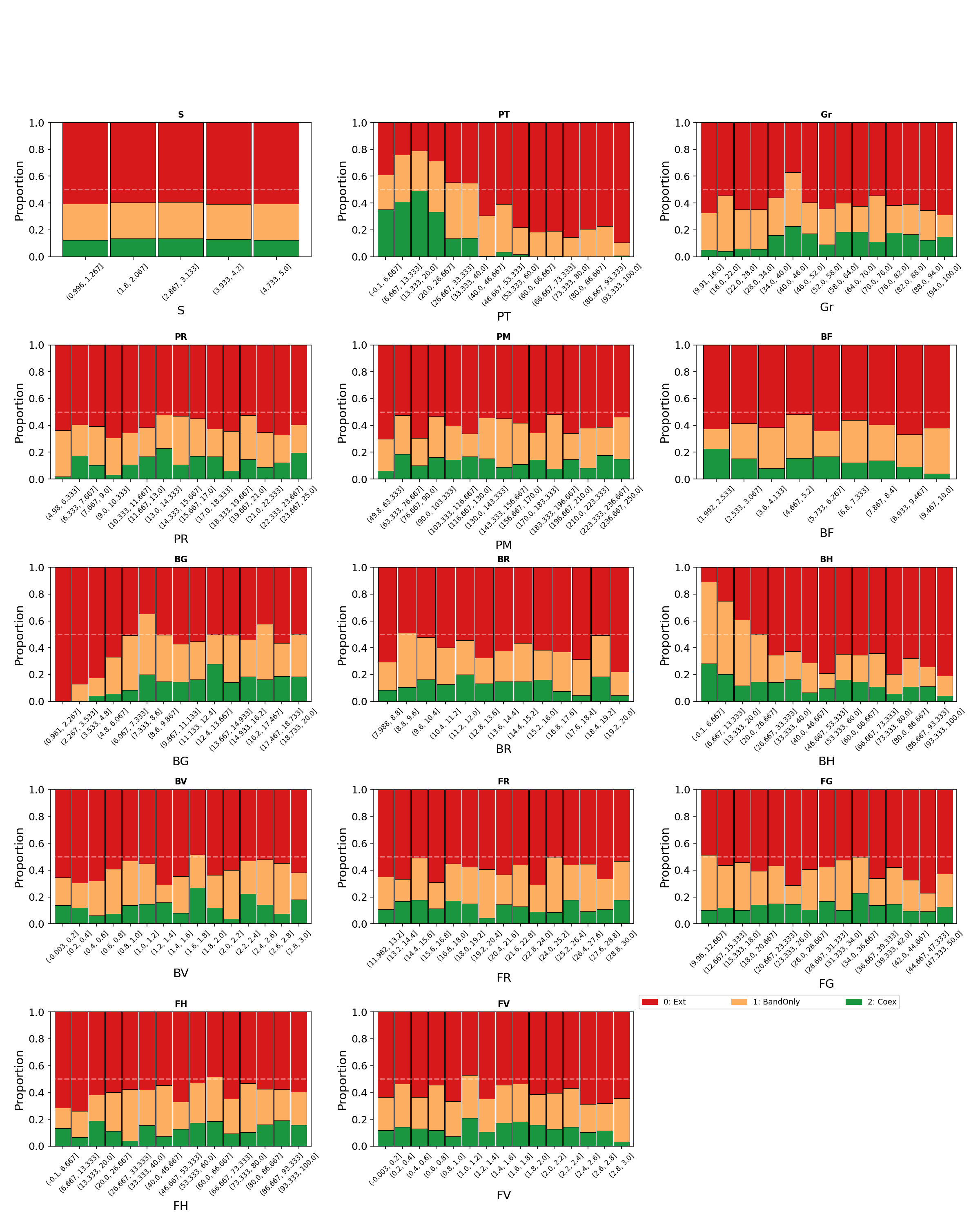}
        \caption{Initial Exploratory Batch ($N=3,250$)}
        \label{fig:regime_v1}
    \end{subfigure}
    \hfill
    \begin{subfigure}[b]{0.49\textwidth}
        \centering
        \includegraphics[width=\textwidth]{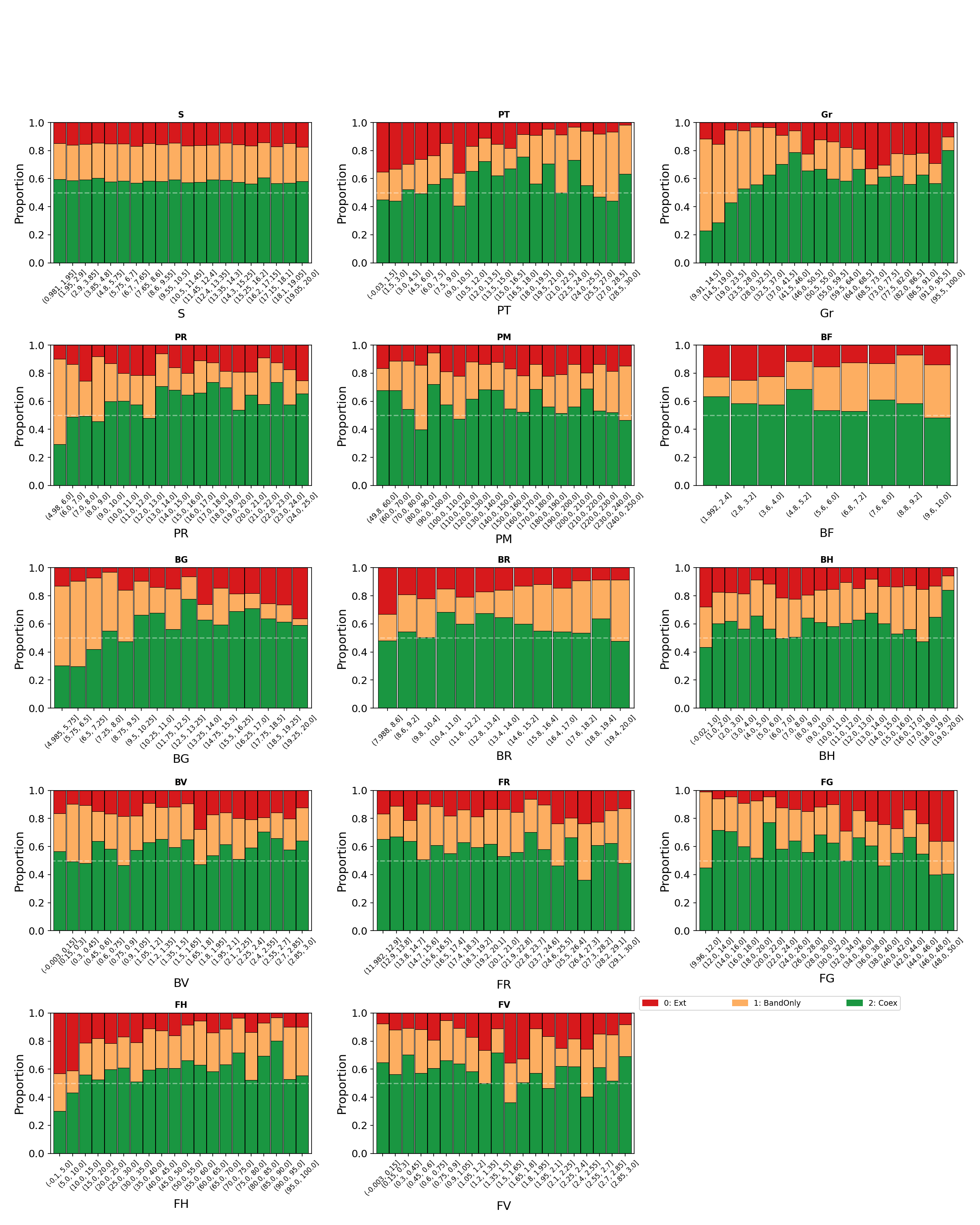}
        \caption{Refined Focused Batch ($N=13,000$)}
        \label{fig:regime_v2}
    \end{subfigure}
    \caption{\textbf{Global Regime Dynamics.} Comparison of simulation outcome distributions between the initial broad exploration (V1) and the refined sampling strategy (V2). The dominance of the Extinction regime highlights the system's structural vulnerability.}
    \label{fig:supp_regimes}
\end{figure}

\begin{figure}[H]
    \centering
    \includegraphics[width=1\textwidth]{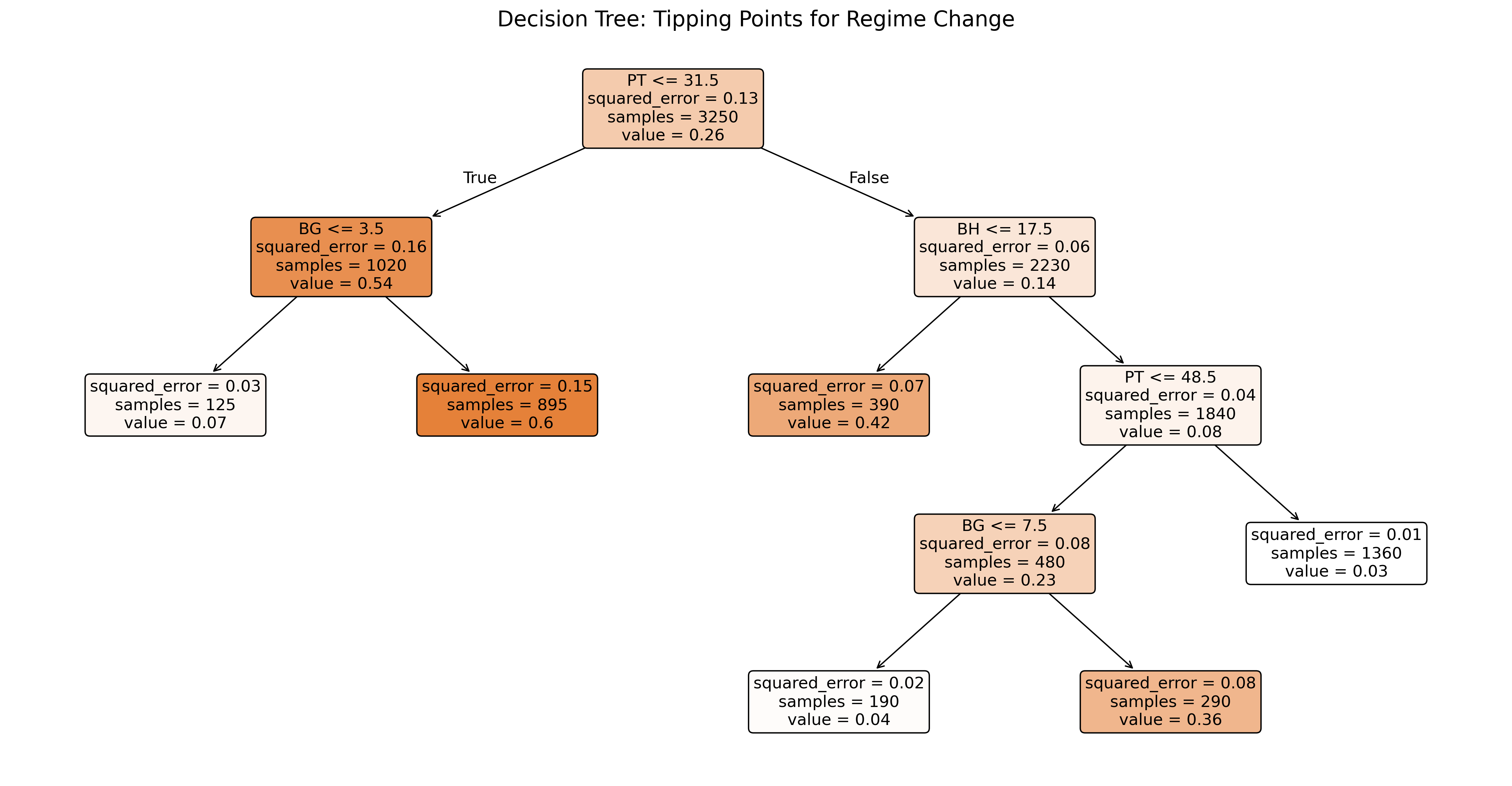}
    \caption{\textbf{Model-Based Screening (Phase 1).} A Decision Tree (CART) trained on the initial dataset. It identifies the primary anthropogenic tipping point: a deterministic shift toward extinction when the Proportion of Hunting Zones ($PH$) exceeds $\approx 31\%$.}
    \label{fig:supp_dt}
\end{figure}

\begin{figure}[H]
    \centering
    \begin{subfigure}[b]{0.48\textwidth}
        \centering
        \includegraphics[width=\textwidth]{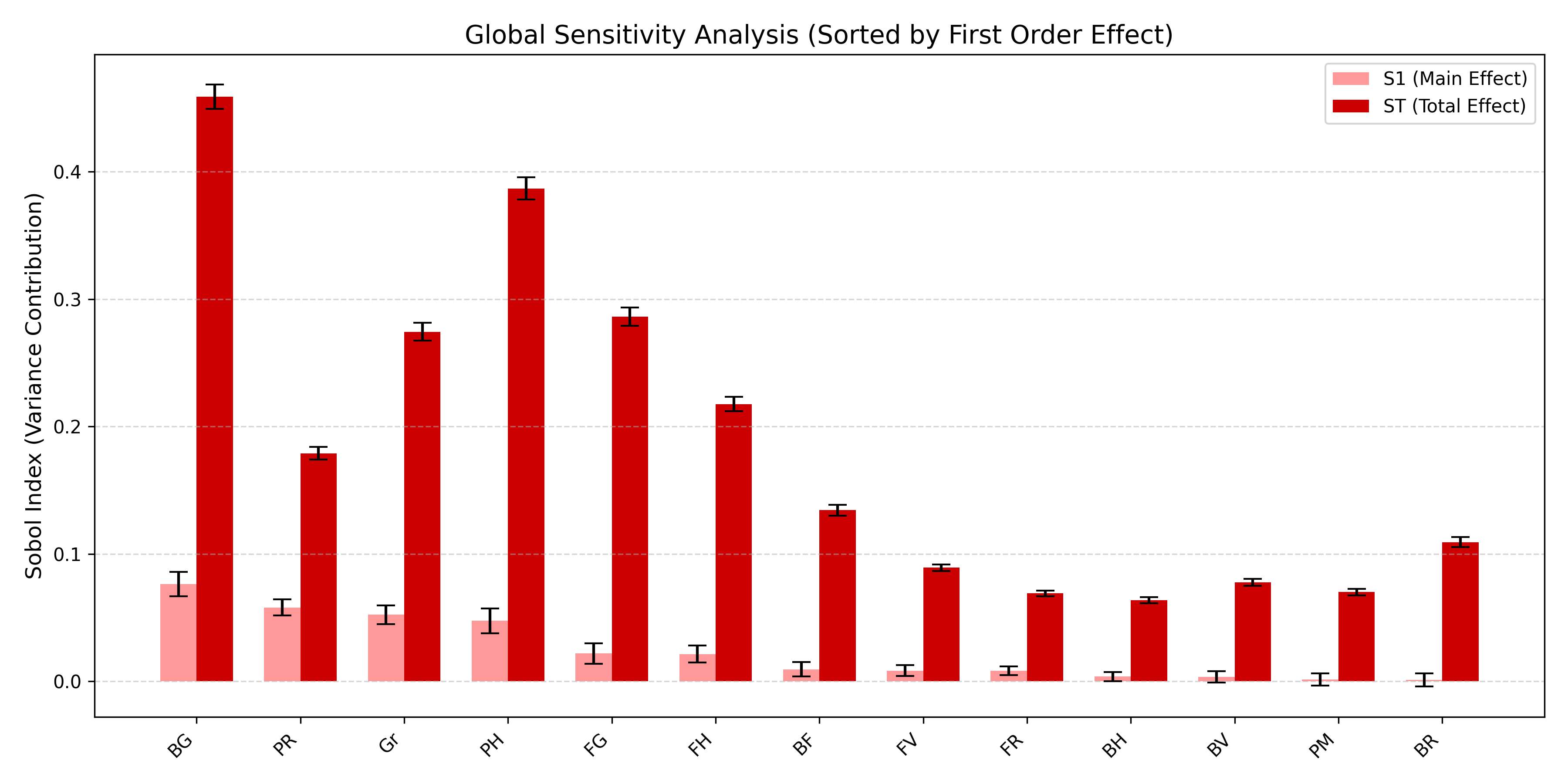}
        \caption{Sobol' Indices ($S_1$ vs $S_T$)}
        \label{fig:supp_sobol_bars}
    \end{subfigure}
    \hfill
    \begin{subfigure}[b]{0.48\textwidth}
        \centering
        \includegraphics[width=\textwidth]{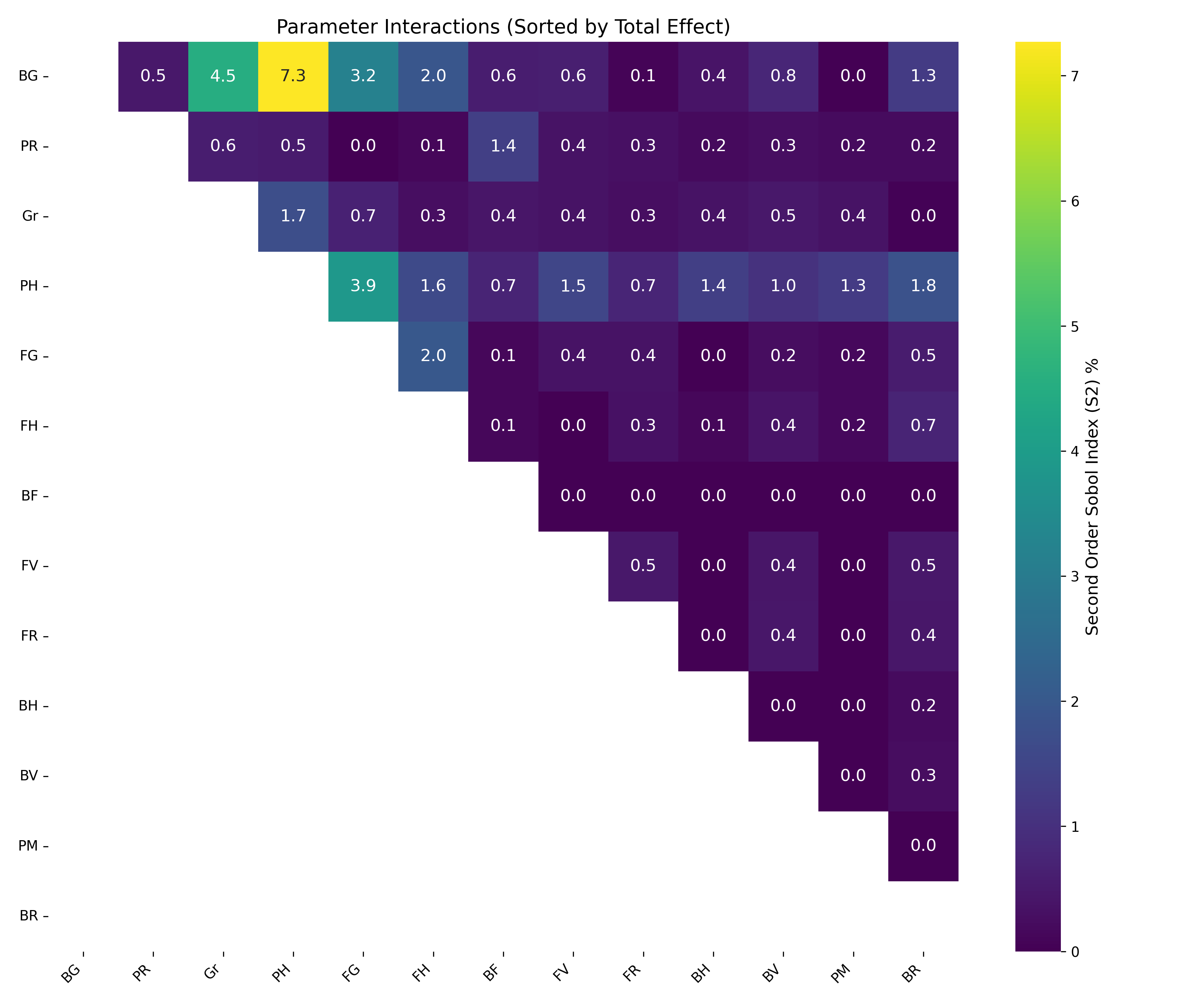}
        \caption{Interaction Heatmap}
        \label{fig:supp_sobol_heat}
    \end{subfigure}
    \caption{\textbf{Global Sensitivity Analysis (Phase 2).} (a) Comparison of First-Order and Total-Order indices reveals that variance is driven by interactions rather than single parameters. (b) The heatmap highlights the strong coupling between Bandicoot Energy Gain ($BG$) and Grassland Availability ($Gr$).}
    \label{fig:supp_sobol}
\end{figure}

\begin{figure}[H]
    \centering
    \includegraphics[width=\textwidth]{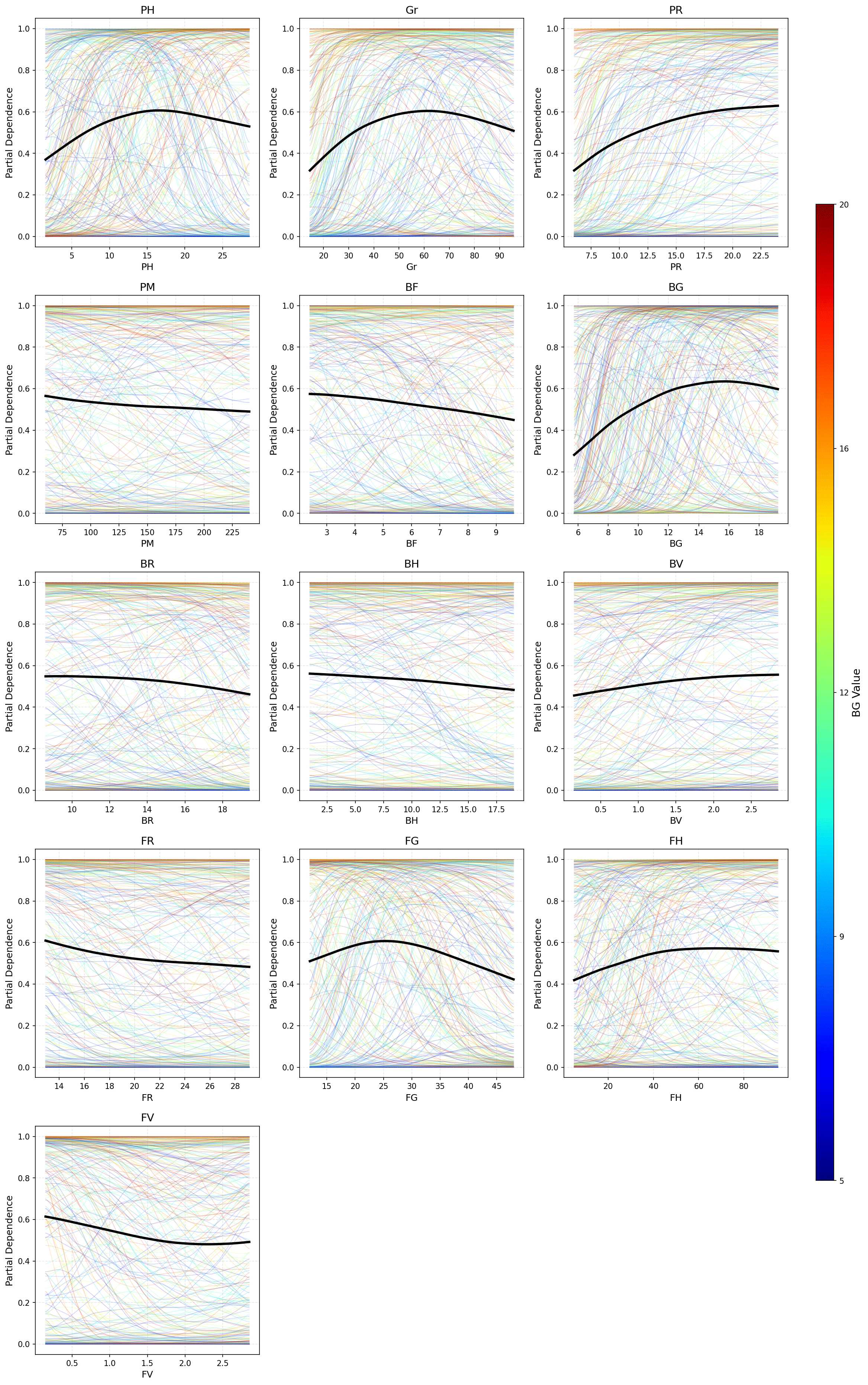}
    \caption{\textbf{Nonlinear Response and Mechanisms.} Partial Dependence Plots (black lines) overlaid with Individual Conditional Expectation curves (colored lines). The coloring by $BG$ reveals a ``Metabolic Trap'': increasing resources ($Gr$) only promotes coexistence if metabolic efficiency ($BG$) is sufficiently high (red lines).}
    \label{fig:supp_pdp}
\end{figure}

\begin{figure}[H]
    \centering
    \includegraphics[width=\textwidth]{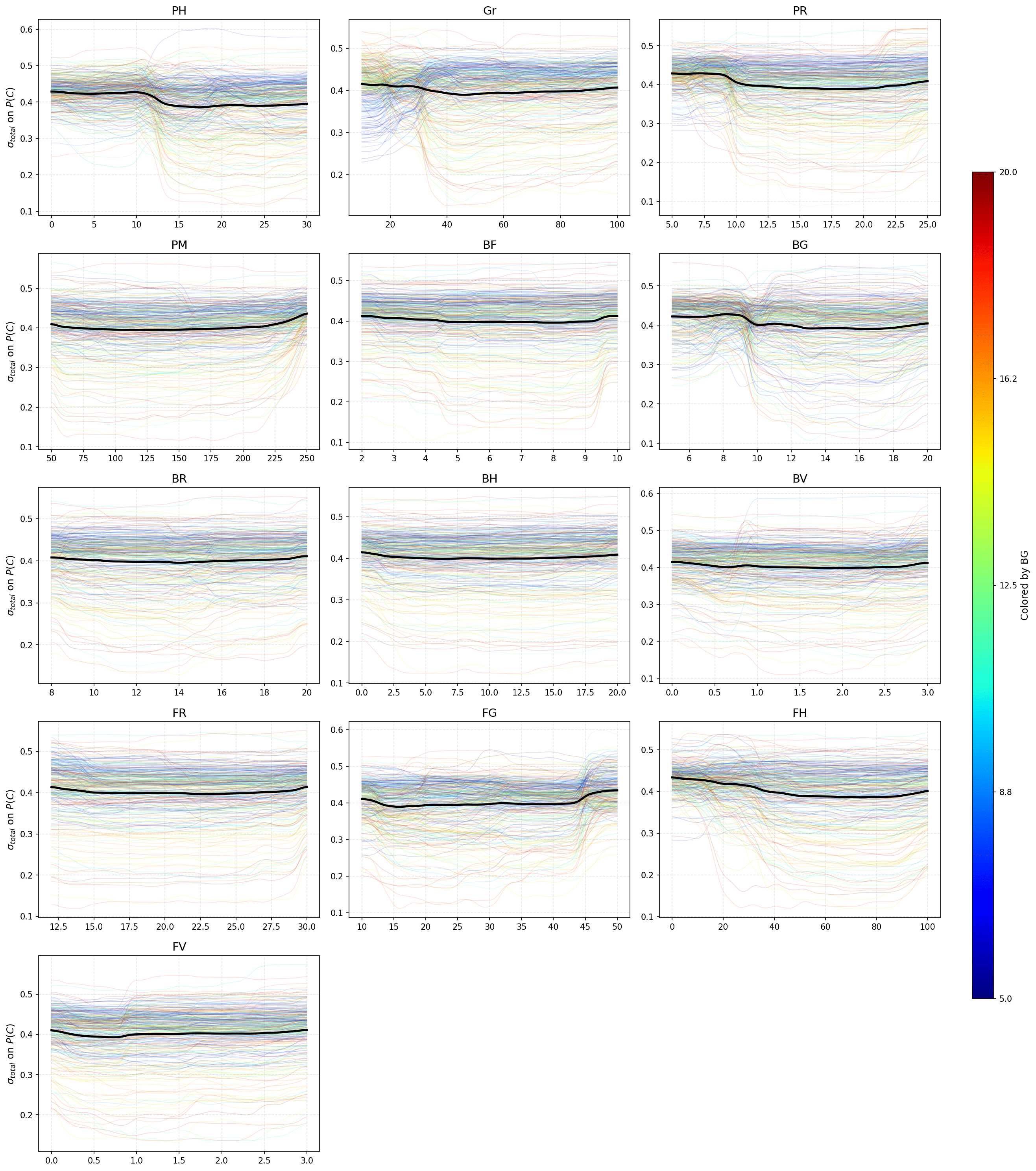}
    \caption{\textbf{Phase Transition Zones.} A map of the Total Euclidean Uncertainty ($\sigma_{total} = \sqrt{\sigma_{aleatoric}^2 + \sigma_{epistemic}^2}$) regarding the probability of coexistence. Peaks in uncertainty identify the precise location of tipping points where the ecosystem is structurally unstable.}
    \label{fig:supp_uncertainty}
\end{figure}

\end{document}